\documentclass{article}
\usepackage{spconf,amsmath,graphicx}
\usepackage{amssymb} 
\usepackage{color}
\usepackage{url}
\usepackage[tight,footnotesize]{subfigure}
\usepackage{multirow}

\title{SUMNet: Fully Convolutional Model for Fast Segmentation of Anatomical Structures in Ultrasound Volumes}
\name{Sumanth Nandamuri, Debarghya China, Pabitra Mitra, Debdoot Sheet}
\address{Indian Institute of Technology Kharagpur}
\begin{document}
%
\maketitle
\begin{abstract}

Ultrasound imaging is generally employed for real-time investigation of internal anatomy of the human body for disease identification. Delineation of the anatomical boundary of organs and pathological lesions is quite challenging due to the stochastic nature of speckle intensity in the images, which also introduces visual fatigue for the observer. This paper introduces a fully convolutional neural network based method to segment organ and pathologies in ultrasound volume by learning the spatial-relationship between closely related classes in the presence of stochastically varying speckle intensity. We propose a convolutional encoder-decoder like framework with (i) feature concatenation across matched layers in encoder and decoder and (ii) index passing based unpooling at the decoder for semantic segmentation of ultrasound volumes. We have experimentally evaluated the performance on publicly available datasets consisting of $10$ intravascular ultrasound pullback acquired at $20$ MHz and $16$ freehand thyroid ultrasound volumes acquired $11 - 16$ MHz. We have obtained a dice score of $0.93 \pm 0.08$ and $0.92 \pm 0.06$ respectively, following a $10$-fold cross-validation experiment while processing frame of $256 \times 384$ pixel in $0.035$s and a volume of $256 \times 384 \times 384$ voxel in $13.44$s.

\end{abstract}
\section{Introduction}

Ultrasonic imaging relies on the principle of acoustic energy propagating through tissues that is partially absorbed, attenuated or backscattered to a varying extent based on the nature of different tissues. The backscattered signal is received by the transducer and processed to render a B-mode image. Radiologist's infer the anatomy of an organ and pathology of a lesion by visually observing the stochastic characteristics of speckles. However, co-located heterogeneous tissue composition often results in non-unique stochastic patterns in speckle intensity. This is derived as a result of mixing of the backscattered signals, thereby challenging the visual reader's ability. The challenges associated with reporting ultrasound images are primarily on account of (i) the stochastic nature of speckle intensity and the low signal-to-noise ratio leading to lowered contrast between structures~\cite{narayan2017speckle}, (ii) imaging and volume rendering artifacts introduced due to breathing and other kinds of body motion during imaging~\cite{sheet2014joint}, (iii) motion induced artifacts during freehand scanning of 2D frames that are subsequently registered and stacked to form 3D volume~\cite{wunderling2017comparison}.

\begin{figure}[t]
\centerline{\includegraphics[width=0.45\textwidth]{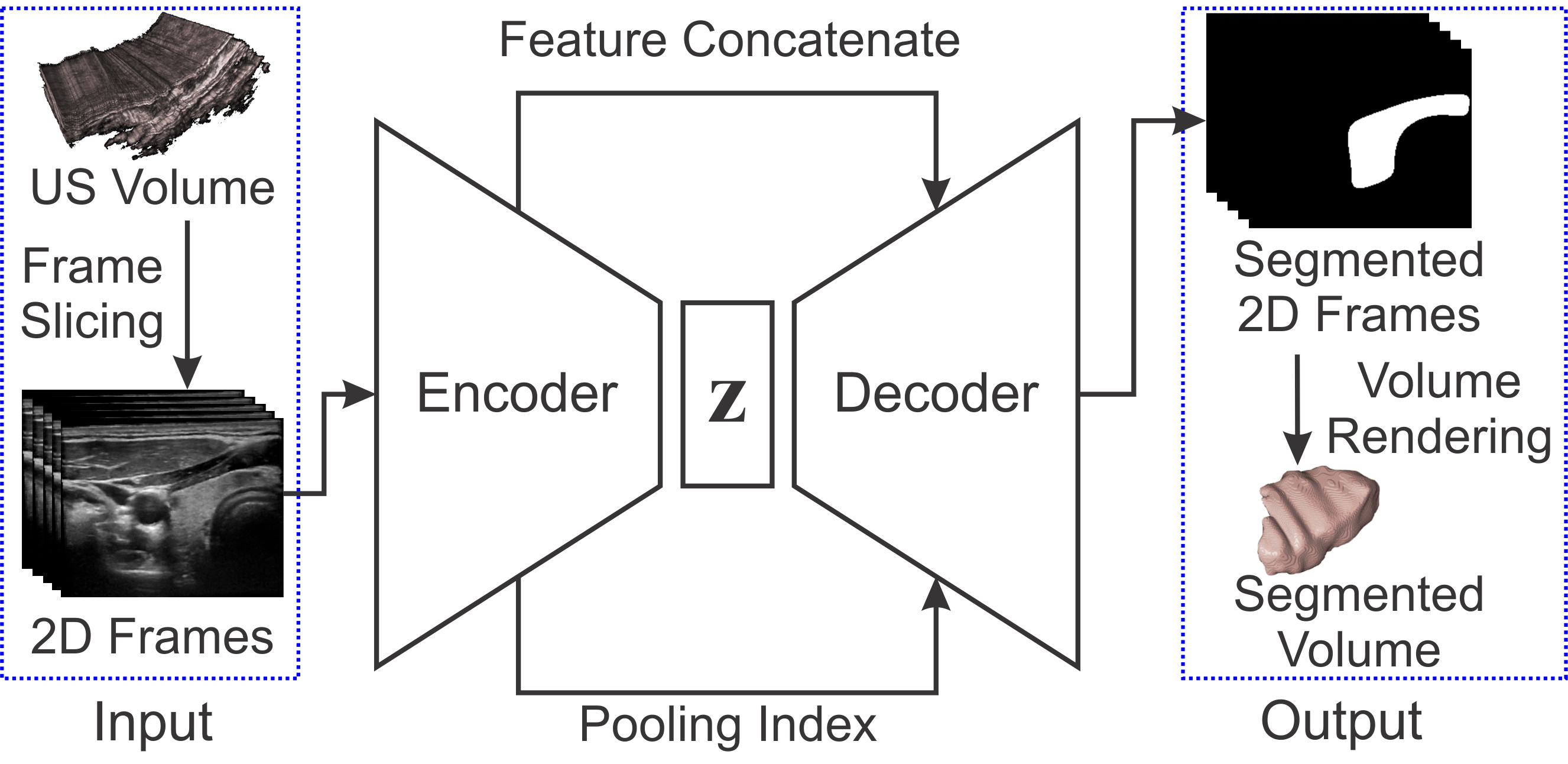}}
\caption{Framework of our method for volumetric segmentation in ultrasound employing a frame level processing.}
\label{fig:SUMFlow}
\end{figure}

In this paper, we have presented a fully convolutional neural network based method for segmentation of anatomical structures in ultrasound volumes using a frame-based semantic segmentation approach. We introduce symbiosis of aspects of two popular deep learning based semantic segmentation networks in the framework illustrated in Fig.~\ref{fig:SUMFlow}. It features (i) a SegNet~\cite{badrinarayanan2017segnet} like encoder-decoder type model with pooling index transfer from encoder for upsampling at the decoder at matched depth, (ii) encoder initialization with weights obtained from an ImageNet pre-trained VGG11~\cite{simonyan2014very}, and (iii) activation transfer from encoder to decoder for concatenation as featured in U-Net~\cite{ronneberger2015u}. These features together help us build a model combining the strengths of the computationally efficient semantic segmentation networks for natural images~\cite{badrinarayanan2017segnet}, along with the one for biomedical images~\cite{ronneberger2015u}. Furthermore, we employ a weighted loss function similar to \cite{ronneberger2015u} during the learning process to preserve the contour.

\begin{figure*}[t]
\centerline{\includegraphics[width=0.9\textwidth]{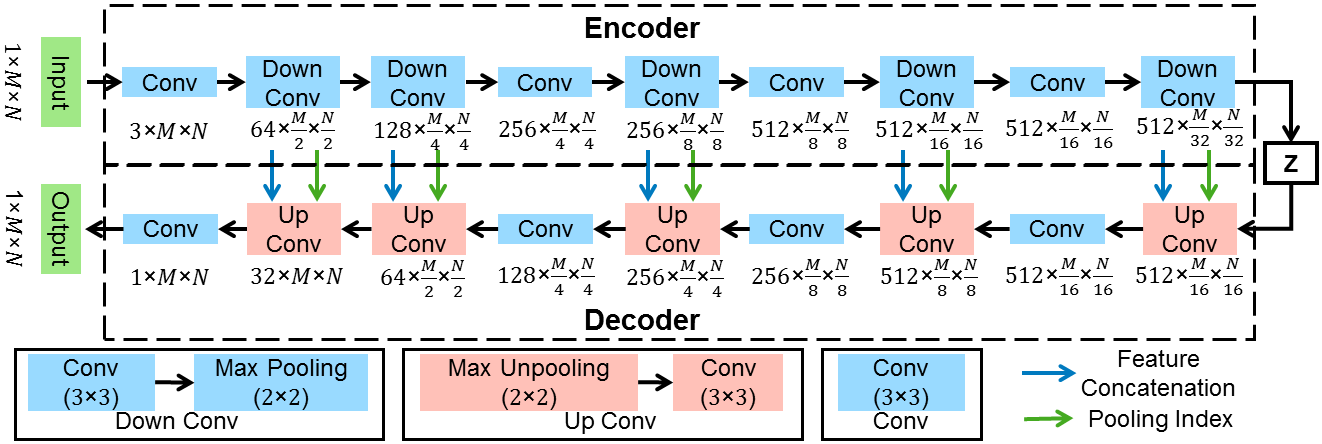}}
\caption{Detailed architecture of the SUMNet for ultrasound volume segmentation.}
\label{SUMFlow}
\end{figure*}

This paper is organized as follows. The existing techniques for ultrasound segmentation are briefly described in Sec.~\ref{sec:Prior Art}. The exposition to the solution is detailed in Sec.~\ref{Exposition to the Solution}. The description of data used, experiment performed, results obtained, and their discussions are elucidated in Sec.~\ref{sec:Experiments and Results} and the work is concluded in Sec.~\ref{sec:Conclusions}.

\section{Prior Art}
\label{sec:Prior Art}

We present a comprehensive review of the prior art for segmentation in two widely used imaging methods namely (i) inside-out imaging such as with intravascular ultrasound (IVUS) and (ii) outside-in imaging such as for the thyroid.

\textbf{IVUS Segmentation:} Earlier approaches primarily employed non neural network based approaches such as those based on minimization of the probabilistic cost function~\cite{mendizabal2013segmentation}. This method employed a support vector machine (SVM) trained on the samples of the first frame in a pullback to segment lumen. Balocco \emph{et al.}~\cite{balocco2014standardized} reported a comprehensive evaluation of multiple approaches through the 2011 MICCAI workshop, where the methods were tested on a common data. In an earlier work~\cite{china2018fly} making use of backscattering physics based learning model inspired from ~\cite{sheet2014joint}, we had proposed an automated algorithm for segmentation of lumen and external elastic laminae which was subsequently extended to use in both outside-in and inside out imaging in ~\cite{china2018anatomical}. A fully convolutional network-based architecture, named IVUS-Net~\cite{yang2018ivus} is proposed for semantic segmentation of IVUS. Another fully convolutional neural network based architecture based on modified UNet~\cite{kim2018fully} is proposed for semantic segmentation of IVUS and demonstrated only using $40$ MHz data.

\textbf{Thyroid Segmentation:} Prior art included use of level-set active contours model~\cite{savelonas2009active} capable of handling segmentation of the thyroid US image. An approach to estimate and classify the thyroid gland in blocks was presented in ~\cite{chang2010thyroid}. The main advantage of those models are to reduce the effect of intensity inhomogeneity. Later on, feedforward neural networks have been used for segmentation of thyroid~\cite{narayan2017speckle}. In ~\cite{wunderling2017comparison}, three semi-automated approaches based on the level set, graph cut, and feature classification for thyroid segmentation were proposed, while in ~\cite{narayan2017speckle} authors introduced a multi-organ method for segmenting thyroid gland, carotid artery, muscles, and the trachea.

These algorithms can be grouped as being semi-automatic where they need some basic information about the organ contour, or they employ a prior knowledge about the speckle pattern of the region of interest and the geometric structure of the organ. The algorithms fail to detect the organ boundary, once image quality degrades and contour size decrease. The predominant challenge is that the processing time consumed by these algorithms remains significantly high. Most of the algorithms are not currently usable for 3D volumetric analysis. The motivation of this paper is to resolve the above-mentioned limitations using a deep fully convolution neural network (FCN) for semantic segmentation which we demonstrate here for both inside-out and outside-in ultrasound imaging.

\begin{table*}[t]
\centering
\caption{Performance evaluation in comparison with prior art. EEL denotes \emph{external elastic luminae}, JCC denotes Jaccard Coeff. and higher value is better, HD denotes Hausdorff Distance and lower value is better, and PAD denotes Percentage Area Difference and lower value is better.}
\scriptsize
\label{Table for dataset B}
\renewcommand{\arraystretch}{1}
\linespread{0.05}\selectfont\centering
\begin{tabular}{cccccccccc}
\hline \hline \\
\multirow{2}{*}{\textbf{Methods}} & \multicolumn{2}{c}{\textbf{JCC}}                 & \multicolumn{2}{c}{\textbf{HD}}                  & \multicolumn{2}{c}{\textbf{PAD}}                \vspace{0.05cm} \\ 
                                  & \textit{\textbf{Lumen}} & \textit{\textbf{EEL}} & \textit{\textbf{Lumen}} & \textit{\textbf{EEL}} & \textit{\textbf{Lumen}} & \textit{\textbf{EEL}}  \\ 
\hline \\
{P1 - P8}~\cite{balocco2014standardized}             &  $0.88 \pm 0.05$   &  $0.91 \pm 0.04$  &  $0.34 \pm 0.14$  &  $0.31 \pm 0.12$  &  $0.06 \pm 0.05$   &  $0.05 \pm 0.04$ \\ 
{IVUS-Net}~\cite{yang2018ivus}                       &  $0.90 \pm 0.06$   &  $0.86 \pm 0.11$  &  $0.26 \pm 0.25$  &  $0.48 \pm 0.44$  &  $-$   &  $-$ \\ 
{2D SegNet}~\cite{badrinarayanan2017segnet}              &  $0.93 \pm 0.05$   &  $0.89 \pm 0.03$  &  $0.20 \pm 0.13$  &  $0.33 \pm 0.10$  &  $0.07 \pm 0.06$   &  $0.10 \pm 0.06$ \\ 
{2D UNet}~\cite{ronneberger2015u}                 &  $0.91 \pm 0.06$   &  $0.88 \pm 0.08$  &  $0.23 \pm 0.19$  &  $0.47 \pm 0.31$  &  $0.10 \pm 0.09$  &  $0.17 \pm 0.12$  \\ 
{3D UNet}~\cite{cciccek20163d}                &  $0.73 \pm 0.13$   &  $0.76 \pm 0.09$  &  $0.33 \pm 0.17$  &  $0.34 \pm 0.14$  &  $0.21 \pm 0.15$   &  $0.27 \pm 0.11$ \\ 
\textbf{SUMNet}                    & {  $\mathbf{0.95} \pm \mathbf{0.03}$ } & {  $\mathbf{0.97} \pm \mathbf{0.01}$ } & {  $\mathbf{0.17} \pm \mathbf{0.07}$ } & {  $\mathbf{0.16} \pm \mathbf{0.09}$ } & {  $\mathbf{0.01} \pm \mathbf{0.01}$ } & {  $\mathbf{0.01} \pm \mathbf{0.01}$ } \\ \\ \hline  \hline
\end{tabular}
\end{table*}

\section{Method}
\label{Exposition to the Solution}

The architecture of the FCN used is illustrated in Fig.~\ref{SUMFlow}. It consists of an encoder unit, consisting of convolution and maxpooling layers and a decoder unit consisting of convolution and unpooling layers. The encoder resembles a VGG11~\cite{simonyan2014very} barring the terminal classification layer. Feature maps are downsampled, and the number of features maps increases progressively with layer. The decoder is built to match activation map sizes in layers matched to depth in encoder with use of pooling index transferred for unpooling operation. This improves the boundary segmentation preserving conformity around small structures which is similar to the SegNet~\cite{badrinarayanan2017segnet}. However, max pooling in encoder with a SegNet typically leads to the loss in subtle information related to smaller structures which progressively vanishes across successive depths. To address these limitations, we introduce the activation concatenation concept similar to that used in the UNet~\cite{ronneberger2015u} which is of relevance to biomedical image segmentation tasks. We have employed ReLU~\cite{goodfellow2016deep} as an activation function in all layer except for the terminal layer of decoder where we have used sigmoid activation function~\cite{goodfellow2016deep} to force the output towards the [0,1] extremity. The network is trained with weighted binary cross-entropy (WCE)~\cite{goodfellow2016deep} loss function. Weights are calculated using a morphological distance transform giving higher weights to the pixels closer to the contour as illustrated for the IVUS in Fig.~\ref{IW} and Fig.~\ref{TW} for thyroid. The idea of using both maxpool indices and concatenating techniques helps in improving the preciseness of localization. The input is a grayscale image, and the first layer of the encoder provides an output with $3$ channel feature map to match input size expected in a VGG11.

\section{Experiments, Results and Discussions}
\label{sec:Experiments and Results}
A $10$-fold patient wise cross validation has been used with leave-one-patient-out validation. 

\begin{figure*}[h]
\centering
\subfigure[IVUS]{\label{I} {\setlength\fboxsep{0pt}\setlength\fboxrule{0pt}\fbox{\includegraphics[width=0.13\textwidth]{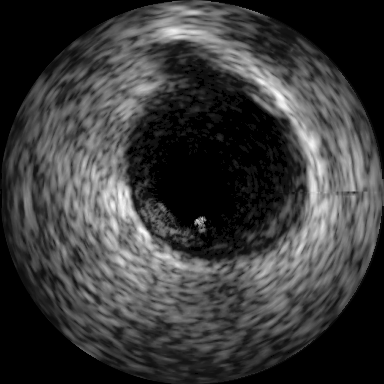}}}}
\subfigure[Ground truth]{\label{IL} {\setlength\fboxsep{0pt}\setlength\fboxrule{0pt}\fbox{\includegraphics[width=0.13\textwidth]{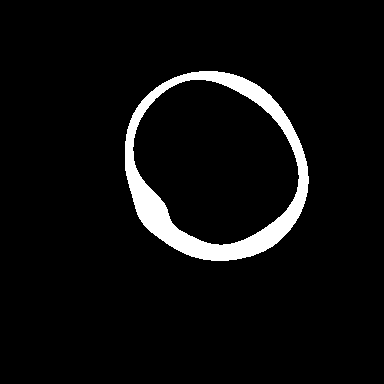}}}}
  \subfigure[Weights]{\label{IW} {\setlength\fboxsep{0pt}\setlength\fboxrule{0pt}\fbox{\includegraphics[width=0.13\textwidth]{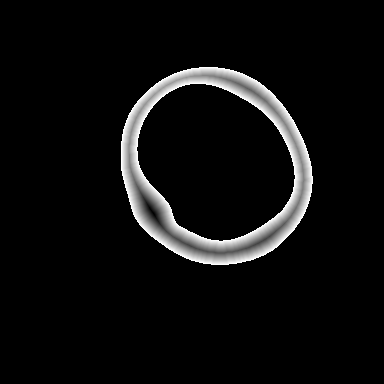}}}}
\subfigure[SUMNet]{\label{fig:IVUS:L} {\setlength\fboxsep{0pt}\setlength\fboxrule{0pt}\fbox{\includegraphics[width=0.13\textwidth]{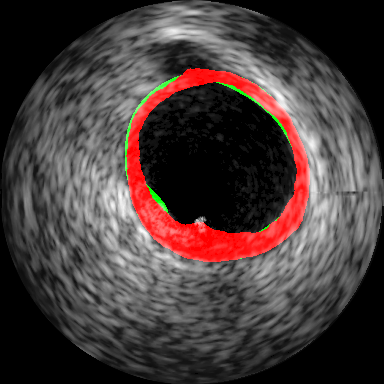}}}}
  \subfigure[2D SegNet]{\label{fig:IVUS:M} {\setlength\fboxsep{0pt}\setlength\fboxrule{0pt}\fbox{\includegraphics[width=0.13\textwidth]{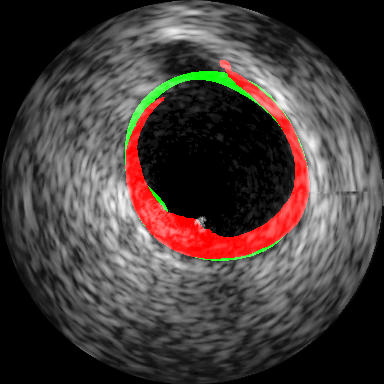}}}}
 \subfigure[2D UNet]{\label{fig:IVUS:GT} {\setlength\fboxsep{0pt}\setlength\fboxrule{0pt}\fbox{\includegraphics[width=0.13\textwidth]{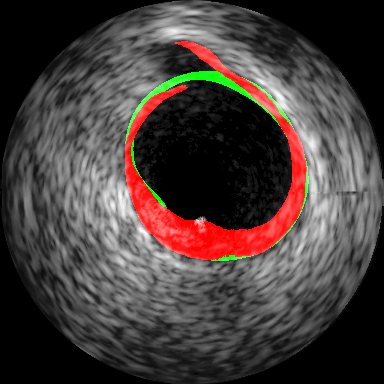}}}}
  \subfigure[3D UNet]{\label{fig:IVUS:Seg} {\setlength\fboxsep{0pt}\setlength\fboxrule{0pt}\fbox{\includegraphics[width=0.13\textwidth]{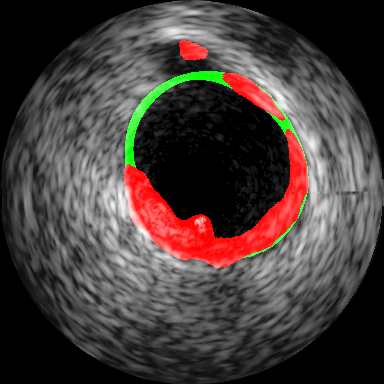}}}}

\subfigure[Thyroid]{\label{T} {\setlength\fboxsep{0pt}\setlength\fboxrule{0pt}\fbox{\includegraphics[width=0.13\textwidth]{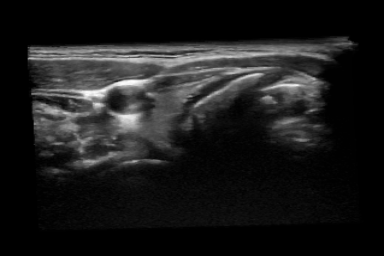}}}}
\subfigure[Ground truth]{\label{TL} {\setlength\fboxsep{0pt}\setlength\fboxrule{0pt}\fbox{\includegraphics[width=0.13\textwidth]{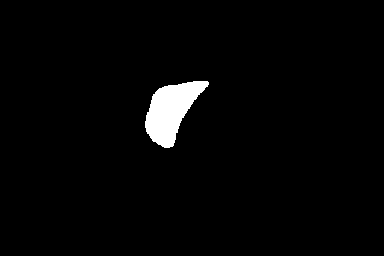}}}}
  \subfigure[Weights]{\label{TW} {\setlength\fboxsep{0pt}\setlength\fboxrule{0pt}\fbox{\includegraphics[width=0.13\textwidth]{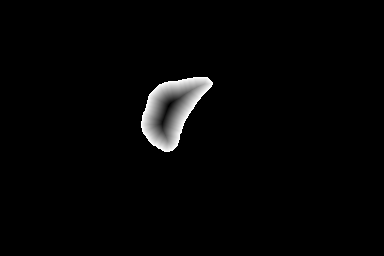}}}}
  \subfigure[SUMNet]{\label{fig:Thyroid:L} {\setlength\fboxsep{0pt}\setlength\fboxrule{0pt}\fbox{\includegraphics[width=0.13\textwidth]{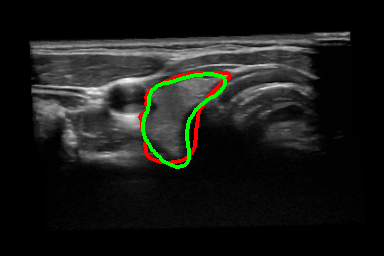}}}}
  \subfigure[2D SegNet]{\label{fig:Thyroid:R} {\setlength\fboxsep{0pt}\setlength\fboxrule{0pt}\fbox{\includegraphics[width=0.13\textwidth]{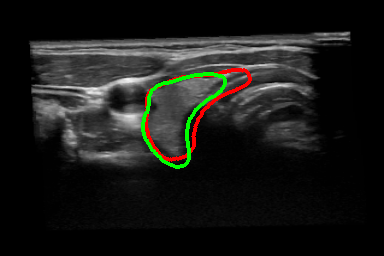}}}}
  \subfigure[2D UNet]{\label{fig:Thyroid:GT} {\setlength\fboxsep{0pt}\setlength\fboxrule{0pt}\fbox{\includegraphics[width=0.13\textwidth]{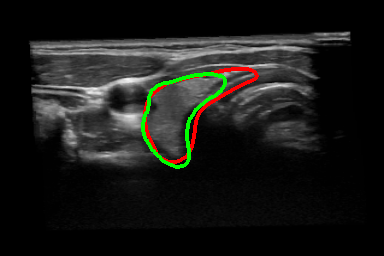}}}}
  \subfigure[3D UNet]{\label{fig:Thyroid:Seg} {\setlength\fboxsep{0pt}\setlength\fboxrule{0pt}\fbox{\includegraphics[width=0.13\textwidth]{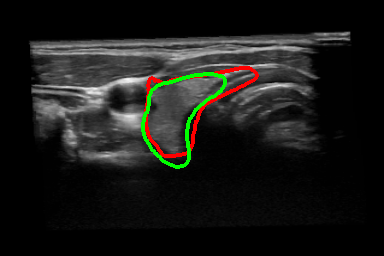}}}}
  
\caption{Weights distribution and 2D visualization of IVUS and thyroid segmentation, where GREEN represents ground truth and RED represents segmentation with the specific method marked in caption.}
\label{fig:3D Vis}
\end{figure*}

\textbf{Intravascular Ultrasound Segmentation:} We have used IVUS data from the 2011 MICCAI workshop~\cite{balocco2014standardized} which is composed of $10$ pullbacks of different patients acquired at $20$ MHz (Dataset B). 
The proposed method is compared with nine different methods, where eight approaches have been reported earlier in~\cite{china2018anatomical}. In Table~\ref{Table for dataset B}, we have included only the best results among the participants (P1-P8) in ~\cite{balocco2014standardized}. Subsequently 2D visualization of segmented IVUS contours are presented in Fig.~\ref{fig:IVUS:L} - (g) and the lumen and external elastic luminae border segmentation in the whole pullback is visualized\footnote[1]{\texttt{Supplementary material}}. The proposed method outperforms the recent related prior art~\cite{yang2018ivus} as presented in Table~\ref{Table for dataset B}.

\begin{table}[h]
\caption{Quantitative analysis and comparison with prior art.}
\scriptsize
\label{thyroid compare}
\begin{tabular}{p{1.7cm}p{1.2275cm}p{1.2275cm}p{1.2275cm}p{1.2275cm}}
\hline
\hline
\textbf{Methods} & \textbf{Sensitivity}   & \textbf{Specificity} & \textbf{Dice} & \textbf{PPV} \\ \hline
Narayan \emph{et al.}~\cite{narayan2017speckle}   & $0.96 \pm 0.03$ & $0.89 \pm 0.07$ & $0.84 \pm 0.05$ & $0.81 \pm 0.09$ \\
JCR~\cite{narayan2017speckle}   & $0.56 \pm 0.07$ & $0.93 \pm 0.07$ & $0.48 \pm 0.07$ & $0.33 \pm 0.06$ \\
Chang \emph{et al.}~\cite{chang2010thyroid}   & $0.87 \pm 0.12$ & $0.56 \pm 0.32$ & $0.51 \pm 0.29$ & $0.53 \pm 0.35$ \\
Garg \emph{et al.}~\cite{narayan2017speckle}& $0.47 \pm 0.18$ & $0.86 \pm 0.22$ & $0.40 \pm 0.14$ & $0.27 \pm 0.11$ \\
2D SegNet~\cite{badrinarayanan2017segnet}   & $0.98 \pm 0.07$ & $0.84 \pm 0.14$ & $0.84 \pm 0.13$ & $0.57 \pm 0.26$ \\ 
2D UNet~\cite{ronneberger2015u}   & $0.98 \pm 0.06$ & $0.87 \pm 0.09$ & $0.86 \pm 0.09$ & $0.63 \pm 0.19$ \\
3D UNet~\cite{cciccek20163d}   & $0.95 \pm 0.15$ & $0.78 \pm 0.30$ & $0.67 \pm 0.11$ & $0.58 \pm 0.14$  \\ 
\textbf{SUMNet}   & ${0.98} \pm {0.01}$ & ${0.89} \pm {0.08}$ & ${0.92} \pm {0.06}$  &  ${0.85} \pm {0.09}$  \\
        \hline
        \hline
    \end{tabular}
\end{table}

\textbf{Thyroid Segmentation:} The thyroid data used in this experiment has been acquired from a publicly available dataset~\cite{wunderling2017comparison} which includes freehand acquired thyroid US volumes from $16$ healthy human subjects imaged with a $11-16$ MHz probe. 
The proposed approach is compared with four different algorithms which has been reported by Narayan et. al.~\cite{narayan2017speckle}, (Table.~\ref{thyroid compare}). The only limitation in trying to compare commonality is that the dataset used for this comparison is not the same as the dataset used in prior art~\cite{narayan2017speckle}. 2D visualization of segmentation is shown in Fig.~\ref{fig:Thyroid:L} - (n) and the thyroid segmentation in the whole volume is visualized~\footnotemark[1]. Table~\ref{thyroid compare} shows that our proposed frame work obtain better results than the prior art.

\textbf{Implementation:} We have experimentally trained the network on a PC with Intel Core i5-8600K CPU, 32GB of system RAM, Nvidia Tesla K40c GPU with 12GB DDR5 RAM. The codes are implemented in Python 3.6 and PyTorch 0.4, Nvidia CUDA 9.0 and cuDNN 5.1 on Ubuntu 16.04 LTS OS. The proposed network is trained with a learning rate of $1 \times 10^{-3}$, and batch size of 14. Adam optimizer~\cite{goodfellow2016deep} has been used to learn the network weights during training. A CPU platform was used for 3D CNN experiments with large scale memory requirements  (Table~\ref{timetable}), consists of $2 \times$ Intel Xeon 8160 CPU with $12 \times 32$GB DDR4 ECC Regd. RAM, Ubuntu 16.04 LTS OS, PyTorch 0.4 on Intel Distribution of Python 3.6 with Intel MKL acceleration.

The proposed network performs better featuring max-unpooling which helps preserve local speckle information, better than transposed convolution where loss of local speckle information due to 4$\times$ padding. 

\begin{table*}[h]
\centering
\caption{Memory and computation benchmarking of different algorithms for Fig~\ref{fig:Thyroid:L}-~\ref{fig:Thyroid:Seg}.}
\scriptsize
\label{timetable}
\begin{tabular}{ccccccccc}
\hline
\hline
\textbf{Method}  & \textbf{Target} & \textbf{Input size} & \textbf{Batch size}    & \textbf{\begin{tabular}[]{@{}c@{}}\# Batches \\ per epoch \end{tabular}} & \textbf{RAM (GB)}             & \textbf{\begin{tabular}[]{@{}c@{}}Training time \\ per epoch (min)\end{tabular}} & \textbf{\begin{tabular}[]{@{}c@{}}Inference time \\ per frame (sec)\end{tabular}}\\ \hline
3D UNet~\cite{cciccek20163d}    & CPU & $256 \times 384 \times 384$ & 1 & 14 & 154  & 29.40 & 0.145  \\
\textbf{SUMNet} & CPU & $256 \times 384$ & 384 & 14  & 148 &   30.45    & 0.070 \\  
 & GPU & $256 \times 384$ & 14 & 384 & 9 &   19.32    & 0.035 \\  
SegNet~\cite{badrinarayanan2017segnet} & CPU & $256 \times 384$ & 384 & 14 & 144 &   26.35    & 0.055 \\ 
 & GPU & $256 \times 384$ & 14 & 384 & 9 &   15.21    & 0.031 \\  
UNet~\cite{ronneberger2015u} & CPU & $256 \times 384$ & 384 & 14 & 145 &   27.26    & 0.068 \\ 
 & GPU & $256 \times 384$ & 14 & 384 & 9 &   18.20    & 0.032 \\ \hline \hline 
\end{tabular}
\end{table*}

\section{Conclusions}
\label{sec:Conclusions}

In this paper, we have presented a frame work for automated segmentation of ultrasound volumes using a convolutional neural network. The performance of the proposed architecture is compared with recent prior art, and it is observed that the proposed method outperforms prior art. This network has an advantages to segment boundary region precisely due to the use of WCE as loss function. Activation concatenation concept similar to UNet has been introduced to overcome the critical information loss which happens due to pooling operation and the network is computationally less expensive.

\bibliographystyle{IEEEbib}
\small
\bibliography{refs}

\end{document}